\documentclass[runningheads]{llncs}
\usepackage{amssymb}
\usepackage{graphicx}
\usepackage{subcaption}
\usepackage[misc]{ifsym}
\begin{document}
\title{Tensor-based Multimodal Learning for Prediction of Pulmonary Arterial Wedge Pressure from Cardiac MRI}

%A Tensor-based Approach for Multimodal Learning from Cardiac MRI to Predict Pulmonary Arterial Wedge Pressure

%
\newcommand{\corrauth}{\textsuperscript{(\Letter)}}
\titlerunning{Multimodal Learning for PAWP Prediction from Cardiac MRI}
% If the paper title is too long for the running head, you can set
% an abbreviated paper title here
%
\author{Prasun C. Tripathi\inst{1}\corrauth \and Mohammod N. I. Suvon\inst{1} \and Lawrence Schobs\inst{1} \and Shuo Zhou\inst{1,2} \and Samer Alabed\inst{3,4,5} \and Andrew J. Swift\inst{3,4,5} \and Haiping Lu\inst{1,2,5}} 

%\author{MNI Suvon}\inst{1,2}
%\author{Lawrence Schob}\inst{3}

\authorrunning{Tripathi et al.}
% First names are abbreviated in the running head.
%If there are more than two authors, 'et al.' is used.

\institute{Department of Computer Science, University of Sheffield, Sheffield, UK \and Centre for Machine Intelligence, University of Sheffield, Sheffield, UK \and Department of Infection, Immunity and Cardiovascular Disease, University of Sheffield, Sheffield, UK \and  Department of Clinical Radiology, Sheffield Teaching Hospitals, Sheffield, UK \and INSIGNEO, Institute for in Silico Medicine, University of Sheffield, Sheffield, UK \\
\email{\{p.c.tripathi\corrauth , m.suvon, laschobs1, shuo.zhou, s.alabed, a.j.swift, h.lu\}@sheffield.ac.uk}}
%\and
%Second Author\inst{2,3}\orcidID{1111-2222-3333-4444} \and
%Third Author\inst{3}\orcidID{2222--3333-4444-5555}}
%
%\authorrunning{F. Author et al.}
% First names are abbreviated in the running head.
% If there are more than two authors, 'et al.' is used.
%
%\institute{Anonymous \and
%Springer Heidelberg, Tiergartenstr. 17, 69121 Heidelberg, Germany
%\email{lncs@springer.com}\\
%\url{http://www.springer.com/gp/computer-science/lncs} \and
%ABC Institute, Rupert-Karls-University Heidelberg, Heidelberg, Germany\\
%\email{\{abc,lncs\}@uni-heidelberg.de}}
%
\maketitle              % typeset the header of the contribution
\begin{abstract}
Heart failure is a severe and life-threatening condition that can lead to elevated pressure in the left ventricle. Pulmonary Arterial Wedge Pressure (PAWP) is an important surrogate marker indicating high pressure in the left ventricle. PAWP is determined by Right Heart Catheterization (RHC) but it is an invasive procedure. A non-invasive method is useful in quickly identifying high-risk patients from a large population. In this work, we develop a tensor learning-based pipeline for identifying PAWP from multimodal cardiac Magnetic Resonance Imaging (MRI). This pipeline extracts spatial and temporal features from high-dimensional scans. For quality control, we incorporate an uncertainty-based binning strategy to identify poor-quality training samples. We leverage complementary information by integrating features from multimodal data: cardiac MRI with short-axis and four-chamber views, and cardiac measurements. The experimental analysis on a large cohort of $1346$ subjects who underwent the RHC procedure for PAWP estimation indicates that the proposed pipeline has a diagnostic value and can produce promising performance with significant improvement over the baseline in clinical practice (i.e., $\Delta$AUC $=0.10$, $\Delta$Accuracy $=0.06$, and $\Delta$MCC $=0.39$). The decision curve analysis further confirms the clinical utility of our method. The source code can be found at: \texttt{https://github.com/prasunc/PAWP}.

\end{abstract}
\section{Introduction}
Heart failure is usually characterized by the inability of the heart to supply enough oxygen and blood to other organs of the body~\cite{emdin2009old}. It is a major cause of mortality and hospitalization ~\cite{savarese2022global}. Elevated Pulmonary Arterial Wedge Pressure (PAWP) is indicative of raised left ventricular filling pressure and reduced contractility of the heart. In the absence of mitral valve or pulmonary vasculature disease, PAWP correlates with the severity of heart failure and risk of hospitalization~\cite{adamson2014wireless}. While PAWP can be measured by invasive and expensive Right Heart Catheterization (RHC), simpler and non-invasive techniques could aid in better monitoring of heart failure patients. Cardiac Magnetic Resonance Imaging (MRI) is an effective tool for identifying various heart conditions and its ability to detect disease and predict outcome has been further improved by machine learning techniques~\cite{assadi2022role}. For instance, Swift~et~al.~\cite{swift2021machine} introduced a machine-learning pipeline for identifying Pulmonary Arterial Hypertension (PAH). Recently, Uthoff~et~al.~\cite{uthoff2020geodesically} developed geodesically smoothed tensor features for predicting mortality in PAH.

Cardiac MRI scans contain high-dimensional spatial and temporal features generated throughout the cardiac cycle. The small number of samples compared to the high-dimensional features poses a challenge for machine learning classifiers. To address this issue, Multilinear Principal Component Analysis (MPCA)~\cite{lu2008mpca} utilizes a tensor-based approach to reduce feature dimensions while preserving the information for each mode, i.e. spatial and temporal information in cardiac MRI. Hence, the MPCA method is well-suited for analyzing cardiac MRI scans. The application of the MPCA method to predict PAWP might further increase the diagnostic yield of cardiac MRI in heart failure patients and help to establish cardiac MRI as a non-invasive alternative to RHC. Existing MPCA-based pipelines for cardiac MRI~\cite{swift2021machine,uthoff2020geodesically,alabed2022machine} rely on manually labeled landmarks that are used for aligning heart regions in cardiac MRI. The manual labeling of landmarks is a cumbersome task for physicians and impractical for analyzing large cohorts. Moreover, even small deviations in the landmark placement may significantly impact the classification performance of automatic pipelines~\cite{schobs2021confidence}. To tackle this challenge, we leverage automated landmarks with uncertainty quantification~\cite{schobs2022uncertainty} in our pipeline. We also extract complementary information from multimodal data from short-axis, four-chamber, and Cardiac Measurements (CM). We use CM features (i.e., left atrial volume and left ventricular mass) identified in the baseline work by Garg~et~al.~\cite{garg2022cardiac} for PAWP prediction.

Our \textbf{main contributions} are summarized as follows:
1) \textbf{Methodology:} We developed a fully automatic pipeline for PAWP prediction using cardiac MRI data, which includes automatic landmark detection with uncertainty quantification, an uncertainty-based binning strategy for training sample selection, tensor feature learning, and multimodal feature integration.
2) \textbf{Effectiveness:} Extensive experiments on the cardiac MRI scans of $1346$ patients with various heart diseases validated our pipeline with a significant improvement ($\Delta$AUC $=0.1027$, $\Delta$Accuracy $=0.0628$, and $\Delta$MCC $=0.3917$) over the current clinical baseline. 
3) \textbf{Clinical utility}: Decision curve analysis indicates the diagnostic value of our pipeline, which can be used in screening high-risk patients from a large population.

\section{Methods}
\label{s2}
\begin{figure}[!t]
\centering
\includegraphics[scale=0.66]{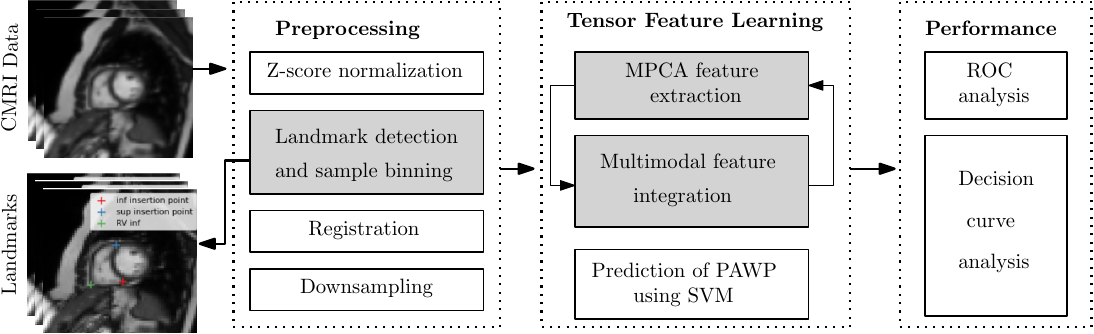}
\caption{The schematic overview of the PAWP prediction pipeline including preprocessing, tensor feature learning, and performance analysis. The blocks in gray color are explained in more detail in Section~\ref{s2}. }
\label{fig1}
\end{figure}

As shown in Fig.~\ref{fig1},  the proposed pipeline for PAWP prediction comprises three components: preprocessing, tensor feature learning, and performance analysis. 
%In this section, we briefly discuss each part of our pipeline.

% \subsubsection{
\vspace{5pt}
\noindent\textbf{Cardiac MRI Preprocessing:}
 The preprocessing of cardiac MRI contains ($1$) normalization of scans, ($2$) automatic landmark detection, ($3$) inter-subject registration, and ($4$) in-plane downsampling. We standardize cardiac MRI intensity levels using Z-score normalization~\cite{jain2005score} to eliminate inter-subject variations. Furthermore, we detect automatic landmarks which is explained in the next paragraph. We perform affine registration to align the heart regions of different subjects to a target image space. We then carry out in-plane scaling of scans by max-pooling at $2$, $4$, $8$, and $16$ times and obtain down-sampled resolutions of $128\times 128$, $64\times 64$, $32\times 32$, and $16\times 16$, respectively.

% \subsubsection{
\vspace{5pt}
\noindent\textbf{Landmark Detection and Uncertainty-based Sample Binning:}
We utilize supervised learning to automate landmark detection using an ensemble of Convolutional Neural Networks (CNNs) for each modality (short-axis and four-chamber). We use the U-Net-like architecture and utilize the same training regime implemented in~\cite{schobs2022uncertainty}. We employ \textit{Ensemble Maximum Heatmap Activation (E-MHA)} strategy~\cite{schobs2022uncertainty} which incorporates an ensemble of five models for each modality. We utilize three landmarks for each modality, with the short-axis modality using the inferior hinge point, superior hinge point, and inferolateral inflection point of the right ventricular apex, and the four-chamber modality using the left ventricular apex and mitral and tricuspid annulus. E-MHA produces an associated uncertainty estimate for each landmark prediction, representing the model's epistemic uncertainty as a continuous scalar value.

A minor error in landmark prediction can result in incorrect image registration~\cite{schobs2021confidence}. To address this issue, we hypothesize that incorrectly preprocessed samples resulting from inaccurate landmarks can introduce ambiguity during model training. For quality control, it is crucial to identify and effectively handle such samples. In this study, we leverage predicted landmarks and epistemic uncertainties to tackle this problem using uncertainty-based binning. To this end, we partition the training scans based on the uncertainty values of the landmarks. The predicted landmarks are divided into $K$ quantiles, i.e., $Q=\{q_{1},q_{2},...,q_{K}\}$, based on the epistemic uncertainty values. We then iteratively filter out training samples starting from the highest uncertain quantile. A sample is discarded if the uncertainty of any of its landmarks lies in quantile $q_{k}$ where $k=\{1,2,..., K\}$. The samples are discarded iteratively until there is no improvement in the validation performance, as measured by the area under the curve (AUC), for two subsequent iterations.

% \subsubsection{
\vspace{5pt}
\noindent\textbf{Tensor Feature Learning:}
To extract features from processed cardiac scans, we employ tensor feature learning, i.e. Multilinear Principal Component Analysis (MPCA) \cite{lu2008mpca}, which learns multilinear bases from cardiac MRI stacks to obtain low-dimensional features for prediction. Suppose we have $M$ scans as third-order tensors in the form of $\{\mathcal{X}_1,\mathcal{X}_2,..,\mathcal{X}_M\in \mathbb{R}^{I_{1} \times I_{2} \times I_{3}}\}$. The low-dimensional tensor features $\{\mathcal{Y}_1,\mathcal{Y}_2,..,\mathcal{Y}_M\in \mathbb{R}^{P_{1} \times P_{2} \times P_{3}}\}$ are extracted by learning three ($N=3$) projection matrices $\{U^{(n)}\in \mathbb{R}^{I_{n} \times P_{n} }, n=1,2,3\}$ as follows:

\begin{equation}
    \mathcal{Y}_m=\mathcal{X}_{m} \times_{1} U^{(1)^{T}} \times_{2} U^{(2)^{T}}\times_{3} U^{(3)^{T}},m=1,2,...,M,
\end{equation}
where $P_{n}<I_{n}$, and $\times_n$ denotes a mode-wise product. Therefore, the feature dimensions are reduced from $I_{1} \times I_{2} \times I_{3}$ to $P_{1} \times P_{2} \times P_{3}$. We optimize the projection matrices $\{U^{(n)}\}$ by maximizing total scatter $\psi_{\mathcal{Y}}=\sum_{m=1}^M||\mathcal{Y}_{m}-\bar{\mathcal{Y}}||_{F}^{2}$, where $\bar{\mathcal{Y}}=\frac{1}{M}\sum_{m=1}^M\mathcal{Y}_{m}$ is the mean tensor feature and $||.||_{F}$ is the Frobenius norm~\cite{lu2013multilinear}. We solve this problem using an iterative projection method. In MPCA,  $\{P_{1},P_{2},P_{3}\}$ can be determined by the explained variance ratio, which is a hyperparameter. Furthermore, we apply Fisher discriminant analysis to select the most significant features based on their Fisher score~\cite{li2018feature}. We select the top $k$-ranked features and employ Support Vector Machine (SVM) for classification.

% \subsubsection{
\vspace{5pt}
\noindent\textbf{Multimodal Feature Integration:}
To enhance performance, we perform multimodal feature integration using features extracted from the short-axis, four-chamber, and Cardiac Measurements (CM). We adopt two strategies for feature integration, namely the early and late fusion of features~\cite{huang2020multimodal}. In early fusion, the features are fused at the input level without doing any transformation. We concatenate features from the short-axis and four-chamber to perform this fusion. We then apply MPCA~\cite{lu2008mpca} on the concatenated tensor, enabling the selection of multimodal features. In late fusion, the integration of features is performed at the common latent space that allows the fusion of features that have different dimensionalities. In this way, we can perform a late fusion of CM features with short-axis and four-chamber features. However, we can not perform an early fusion of CM features with short-axis and four-chamber features.

% \subsubsection{
\vspace{5pt}
\noindent\textbf{Performance Evaluation:}
In this paper, we use three primary metrics: Area Under Curve (AUC), accuracy, and Matthew's Correlation Coefficient (MCC), to evaluate the performance of the proposed pipeline. Decision Curve Analysis (DCA) is also conducted to demonstrate the clinical utility of our methodology.

\begin{table}[!t]
\caption{Baseline characteristics of included patients. $p$ values were obtained using $t$-test~\cite{welch1947generalization}.\label{tab1}}
\centering
\scalebox{0.9}{
\begin{tabular}{l|l|l|l}
\hline
& Low PAWP($\le15$) & High PAWP($>15$) & $p$-value\\
\hline
Number of patients&$940$&$406$& -\\
\hline
Age (in years)&$64.8\pm 14.2$&$70.5\pm 10.6$& $<0.01$\\
\hline
%Male (sex) &$423(28\%)$&$157(10\%)$& $0.06$\\
%\hline
Body Surface Area (BSA) &$1.88\pm 0.28$&$1.93\pm 0.24$& $<0.01$\\
\hline
Heart Rate (bpm) &$73.9\pm 15.5$&$67.6\pm 15.9$& $<0.01$\\
\hline
Left Ventricle Mass (LVM) &$92.3\pm 25$&$106\pm 33.1$& $<0.01$\\
\hline
Left Atrial Volume ($ml^{2}$) &$72.2\pm 33.7$&$132.2\pm 56.7$& $<0.01$\\
\hline
PAWP (mmHg) &$10.3\pm 3.1$&$21.7\pm 4.96$& $<0.01$\\
\hline
\end{tabular}}
\vspace{-10pt}
\end{table}

\section{Experimental Results and Analysis}

\noindent\textbf{Study Population}: 
Patients with suspected pulmonary hypertension were identified after institutional review board approval and ethics committee review. A total of $1346$ patients who underwent Right Heart Catheterization (RHC) and cardiac MRI scans within $24$ hours were included. Of these patients, $940$ had normal PAWP ($\le15$ mmHg), while $406$ had elevated PAWP ($>15$ mmHg). Table~\ref{tab1} summarizes baseline patient characteristics. RHC was performed using a balloon-tipped $7.5$ French thermodilution catheter.

\vspace{5pt}
\noindent\textbf{Cardiac MRI and measurement:}
MRI scans were obtained using a $1.5$ Tesla whole-body GE HDx MRI scanner (GE Healthcare, Milwaukee, USA) equipped with $8$-channel cardiac coils and retrospective electrocardiogram gating. Two cardiac MRI protocols, short-axis and four-chamber, were employed, following standard clinical protocols to acquire cardiac-gated multi-slice steady-state sequences with a slice thickness of $8$ mm, a field of view of $48\times43.2$, a matrix size of $512\times512$, a bandwidth of $125$ kHz, and TR/TE of $3.7/1.6$ ms. Following ~\cite{garg2022cardiac}, left ventricle mass and left atrial volume were selected as cardiac measurements.

\vspace{5pt}
\noindent\textbf{Experimental Design:}
We conducted experiments on short-axis and four-chamber scans across four scales. To determine the optimal parameters, we performed $10$-fold cross-validation on the training set. From MPCA, we selected the top $210$ features. We employed early and late fusion on short-axis and four-chamber scans, respectively, while CM features were only fused using the late fusion strategy.  We divided the data into a training set of $1081$ cases and a testing set of $265$ cases. To simulate a real testing scenario, we designed the experiments such that patients diagnosed in the early years were part of the training set, while patients diagnosed in recent years were part of the testing set. We also partitioned the test into $5$ parts based on the diagnosis time to perform different runs of methods and report standard deviations of methods in comparison results. For SVM, we selected the optimal hyper-parameters from $\{0.001, 0.01, 0.1, 1\}$ using the grid search technique. The code for the experiments has been implemented in Python (version $3.9$). We leveraged the cardiac MRI preprocessing pipeline and MPCA from the Python library PyKale~\cite{lu2022pykale} and SVM implementation is taken from scikit-learn~\cite{scikit-learn}.

\begin{figure}[t]
\centering
\includegraphics[height=3.5cm, width=11cm]{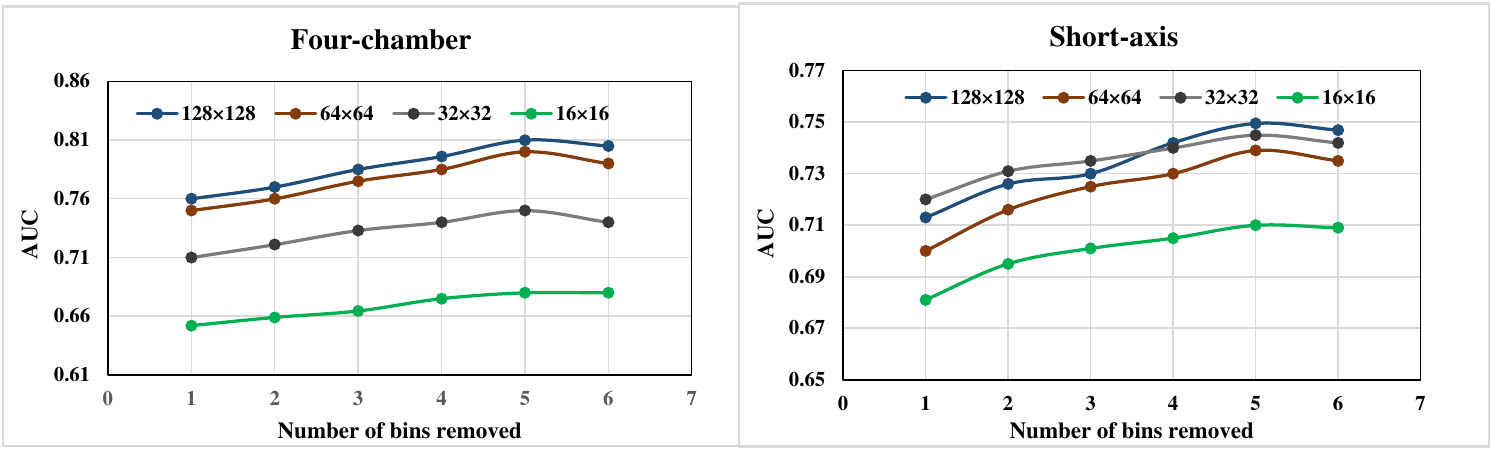}
\caption{Performance comparison of removing a different number of bins of training data on $10$-fold cross-validation. }
\label{binning}

\end{figure}

\begin{figure}[!t]
	\centering
	\begin{subfigure}[b]{\textwidth}
		\centering
		\includegraphics[width=\textwidth]{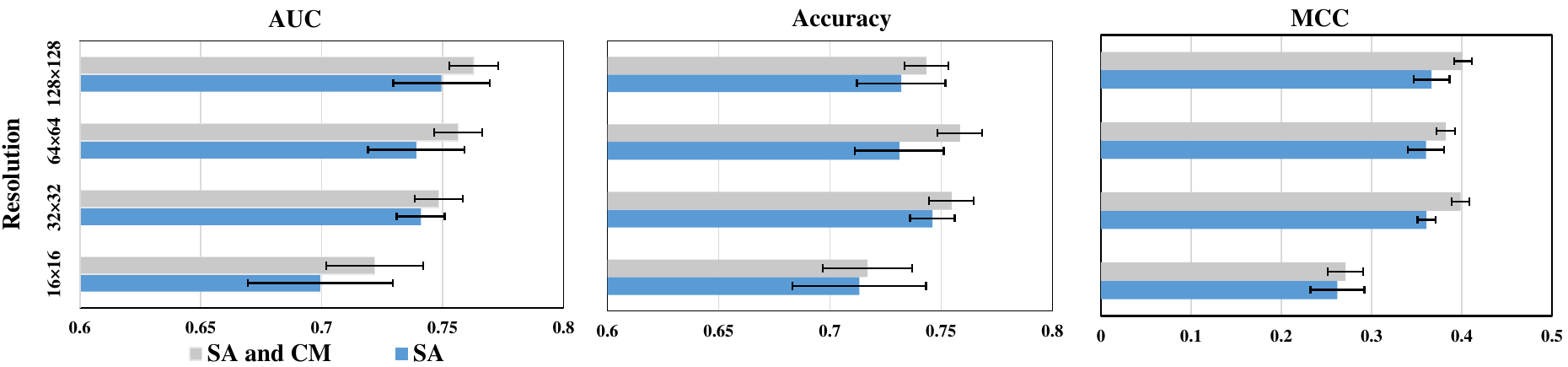}
		\caption{Short-axis versus short-axis and CM}
	\end{subfigure}
 \begin{subfigure}[b]{\textwidth}
		\centering
		\includegraphics[width=\textwidth]{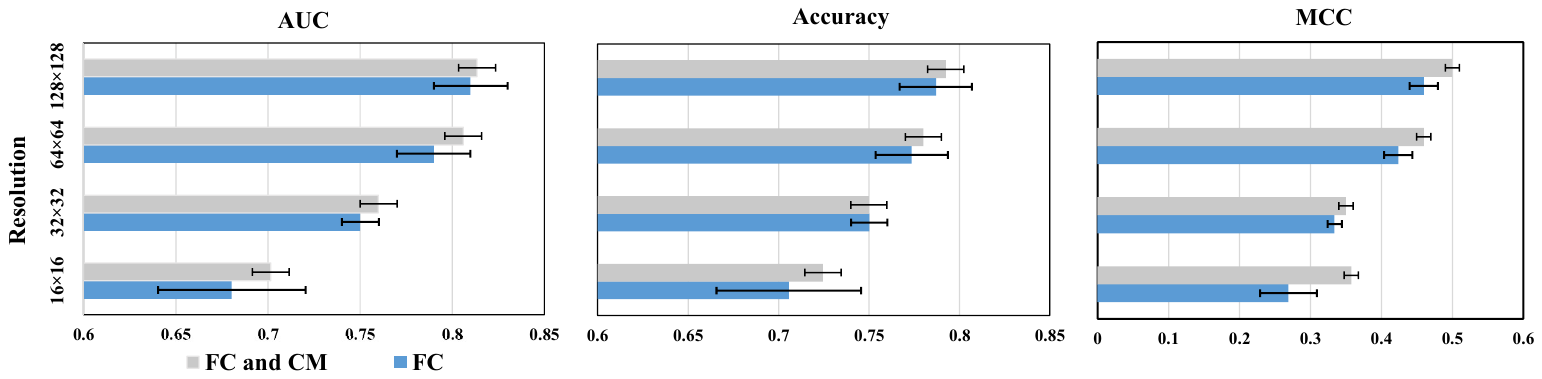}
		\caption{Four-chamber versus four-chamber and CM}
	\end{subfigure}
 
\caption{The effect of combining CM features on short-axis and four-chamber. SA: Short-axis; FC: Four-chamber. }\label{fig2}
\vspace{-10pt}
\end{figure}

\begin{table}[t]
\caption{Performance comparison using three metrics (with \textbf{best} in bold and \underline{second best} underlined). FC: Four-Chamber features; SA: Short-Axis features; CM: Cardiac Measurement features. The standard deviations of methods were obtained by dividing the test set into $5$ parts based on the diagnosis time.   }\label{tab2}
\centering
\scalebox{0.88}{
\begin{tabular}{l|c|c|c|c}
\hline
Modality & Resolution & AUC & Accuracy & MCC\\
\hline
Unimodal (CM)~\cite{garg2022cardiac} &-&$0.7300\pm 0.04$&$0.7400\pm 0.03$&$0.1182 \pm 0.03$\\
\hline
       Unimodal (SA)~\cite{swift2021machine} &$64\times64$&$0.7391\pm 0.05$&$0.7312 \pm 0.07$&$0.3604\pm0.02$ \\
        &$128\times128$&$0.7495\pm 0.05$&$0.7321\pm 0.04$&$0.3277\pm0.01$ \\
        \hline
Unimodal (FC)~\cite{swift2021machine}    &$64\times64$&$0.8034 \pm 0.02$&$0.7509 \pm 0.04$&$0.4240\pm0.02$ \\
        &$128\times128$&$0.8100 \pm 0.04$&$0.7925 \pm 0.05$&$0.4666\pm0.02$ \\
        \hline
  Bi-modal (SA and FC):    &$64\times64$&$0.7998\pm0.01$&$0.7698\pm0.03$&$0.4185\pm0.03$ \\
      Early fusion   &$128\times128$&$0.7470\pm0.02$&$0.7283\pm0.02$&$0.3512\pm0.02$ \\
        \hline
   Bi-modal (SA and FC):   &$64\times64$&$0.8028\pm0.04$&$0.7509\pm0.03$&$0.3644\pm0.01$ \\
    Late fusion     &$128\times128$&$0.8122\pm0.03$&$0.7547\pm0.03$&$0.3594\pm0.02$ \\
        \hline
  Bi-modal (SA and CM):   &$64\times64$&$0.7564\pm0.04$&$0.7585\pm0.02$&$0.3825\pm0.02$ \\
       Late fusion   &$128\times128$&$0.7629\pm0.03$&$0.7434\pm0.03$&$0.3666\pm0.03$ \\
        \hline

Bi-modal (FC and CM):     &$64\times64$&$0.8061 \pm 0.03$&$0.7709 \pm 0.02$&$0.4435 \pm 0.02$ \\
   Late fusion      &$128\times128$&\underline{$0.8135\pm0.02$}&\underline{$0.7925\pm0.02$}&\underline{$0.4999\pm0.03$} \\
        \hline
 Tri-modal (FC, SA, and CM) &$64\times64$&$0.8146 \pm 0.04$&$0.7774 \pm 0.03$&$0.4460 \pm 0.02$ \\
   Hybrid fusion     &$128\times128$&$\mathbf{0.8327\pm0.06}$&$\mathbf{0.8038 \pm 0.05}$&$\mathbf{0.5099\pm0.04}$ \\
\hline
 Tri-modal Hybrid fusion &$64\times64$&$0.7892\pm0.04$&$0.7513 \pm 0.05$&$0.4278\pm0.02$ \\
  without uncertainty binning      &$128\times128$&$0.8036\pm0.03$&$0.7820 \pm 0.04$&$0.4779\pm0.01$ \\
\hline
\end{tabular}}
\vspace{-10pt}
\end{table}

\vspace{10pt}
\noindent\textbf{Uncertainty-Based Sample Binning:}
To improve the quality of training data, we used quantile binning to remove training samples with uncertain landmarks. The landmarks were divided into $50$ bins, and then removed one bin at a time in the descending order of their uncertainties. Figure~\ref{binning} depicts the results of binning using $10$-fold cross-validation on the training set, where  the performance improves consistently over the four scales when removed bins $\le 5$. Based on the results, we removed $5$ bins ($129$ out of $1081$ samples) from the training set, and used the remaining $952$ training samples for the following experiments.

\vspace{5pt}
\noindent\textbf{Unimodal Study:}
The performance of three models on single-modality is reported in Table~\ref{tab2}, including short-axis (SA), four-chamber (FC), and cardiac measurements (CM), where the CM based unimodal is considered as the baseline. The results demonstrate an improvement of $\Delta$AUC $=0.0800$ $\Delta$Accuracy $=0.0527$, and $\Delta$MCC $=0.3484$ over the baseline obtained by FC based unimodal, which indicates that tensor-based features have a diagnostic value.

\vspace{5pt}
\noindent\textbf{Bi-modal Study:}
In this experiment, we compared the performance of bi-modal models. As shown in Table~\ref{tab2}, bimodal (four-chamber and CM) produces superior performance (i.e., AUC $=0.8135$, Accuracy=$0.7925$ and MCC $=0.4999$) among bi-modal models. Next, we investigated the effect of fusing CM features with short-axis and four-chamber modalities in Fig.~\ref{fig2}. It can be observed from these figures that the fusion of CM features enhances the diagnostic power of cardiac MRI modalities at all scales. The bi-modal (four-chamber and CM) model achieved the improvement in the performance ($\Delta$AUC $=0.0035$ and $\Delta$MCC $=0.0333$) over the unimodal (four-chamber) model.

\begin{figure}[!t]
\centering
\includegraphics[height=3.5cm, width=12cm]{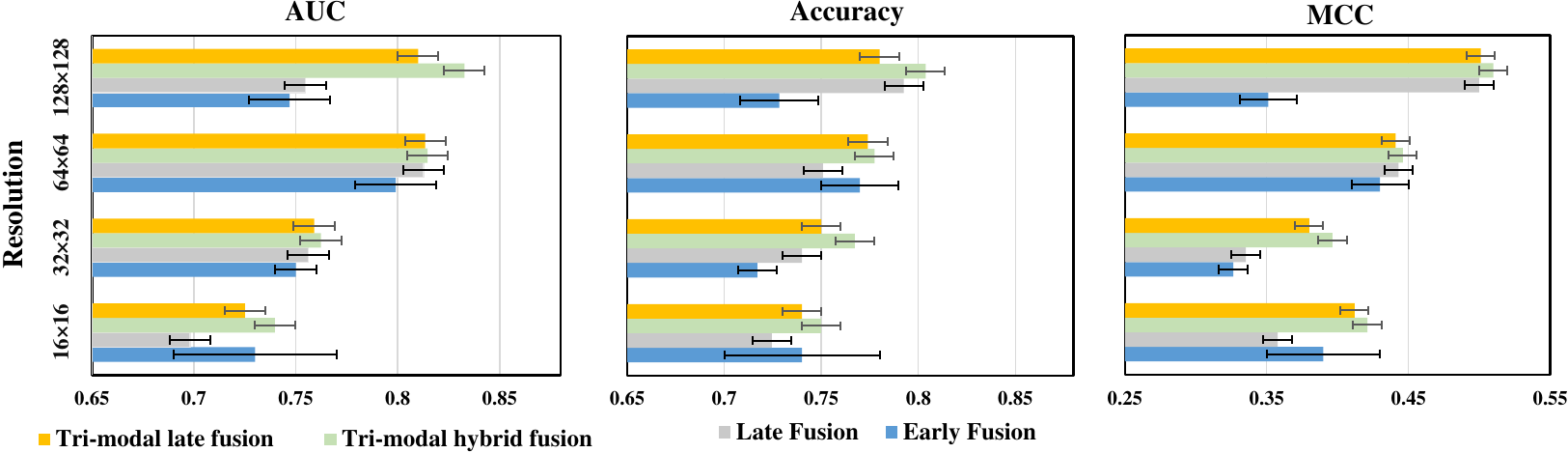}
\caption{The effect of combining CM features on the bi-modals including early and late fusion of four-chamber and short-axis. Early fusion: early fusion of short-axis and four-chamber; late fusion: late fusion of short-axis and four-chamber. }
\label{fig4}
\vspace{-5pt}
\end{figure}

\vspace{5pt}
\noindent\textbf{Effectiveness of Tri-modal:}
In this experiment, we performed a fusion of CM features with the bi-modal models to create two tri-modal models. The first tri-modal is tri-modal late (CM with a late fusion of short-axis and four-chamber) and the second tri-modal is a tri-modal hybrid (CM with an early fusion of short-axis and four-chamber). As shown in Fig.~\ref{fig4}, CM features enhance the performance of bi-modal models and tri-modal hybrid outperforms all. The tri-modal hybrid obtained the best performance (Table~\ref{tab2}, where AUC $=0.8327$, Accuracy $=0.8038$, and MCC $=0.5099$) and a significant improvement of $\Delta$AUC $=0.1027$, $\Delta$Accuracy $=0.0628$, and $\Delta$MCC $=0.3917$ over the baseline method.

\begin{figure}[!t]
\centering
\includegraphics[height=4.5cm, width=6.5cm]{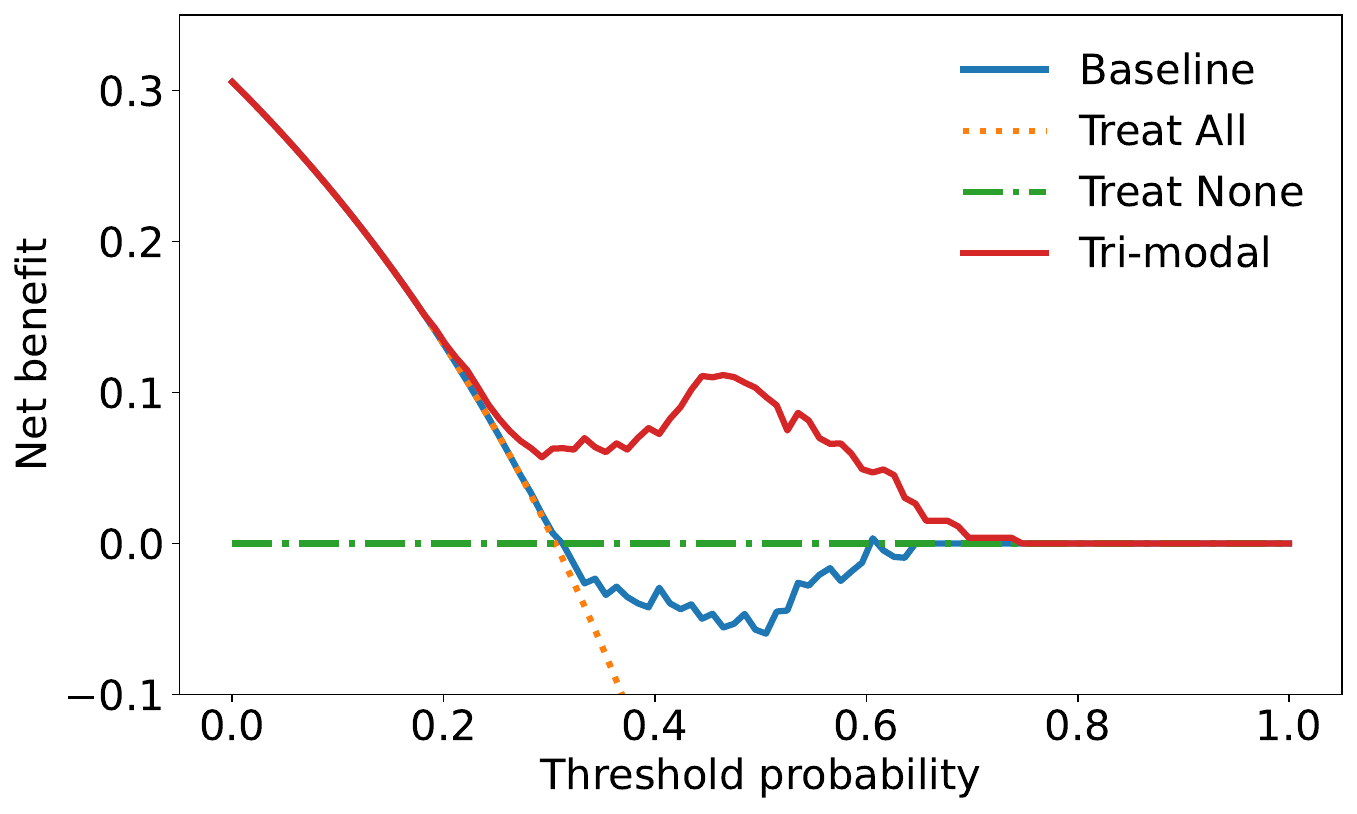}
\caption{Evaluating clinical utility of our method using Decision Curve Analysis (DCA)~\cite{vickers2006decision}.``Treat All" means treating all patients, regardless of their actual disease status, while ``Treat None" means treating no patients at all. Our predictive model's net benefit is compared with the net benefit of treating everyone or no one to determine its overall utility.}
\label{fig5}
\vspace{-10pt}
\end{figure}

\vspace{5pt}
\noindent\textbf{Decision Curve Analysis} (DCA)~\cite{vickers2006decision,sadatsafavi2021moving} on the performance suggests the potential clinical utility of the proposed method. As shown in Fig.~\ref{fig5}, the Tri-modal model outperformed the baseline method for most possible benefit/harm preferences, where benefit indicates a positive net benefit (i.e. correct diagnosis) and harm indicates a negative net benefit (i.e. incorrect diagnosis). The tri-modal model (the best model) obtained a higher net benefit between decision threshold probabilities of $0.30$ and $0.70$ which implies that our method has a diagnostic value and can be used in screening high-risk patients from a large population. 

\vspace{15pt}
\noindent\textbf{Feature contributions:}
Our model is interpretable. The highly-weighted features were detected in the left ventricle and interventricular septum in cardiac MRI. For cardiac measurements, left atrial volume (0.778/1) contributed more than left ventricular mass (0.222/1) to the prediction.

\section{Conclusions}
This paper proposed a tensor learning-based pipeline for PAWP classification. We demonstrated that: $1$) tensor-based features have a diagnostic value for PAWP, $2$) the integration of CM features improved the performance of unimodal and bi-modal methods, $3$) the pipeline can be used to screen a large population, as shown using decision curve analysis. However, the current study is limited to single institutional data. In the future, we would like to explore the applicability of the method for multi-institutional data using domain adaptation techniques.

\section*{Acknowledgment}
The study was supported by the Wellcome Trust grants 215799/Z/19/Z and 205188/Z/16/Z.

\bibliographystyle{splncs04}
\bibliography{mybibliography}

\begin{thebibliography}{10}
\providecommand{\url}[1]{\texttt{#1}}
\providecommand{\urlprefix}{URL }
\providecommand{\doi}[1]{https://doi.org/#1}

\bibitem{adamson2014wireless}
Adamson, P.B., Abraham, W.T., Bourge, R.C., Costanzo, M.R., Hasan, A., Yadav,
  C., Henderson, J., Cowart, P., Stevenson, L.W.: Wireless pulmonary artery
  pressure monitoring guides management to reduce decompensation in heart
  failure with preserved ejection fraction. Circulation: Heart Failure
  \textbf{7}(6),  935--944 (2014)

\bibitem{alabed2022machine}
Alabed, S., Uthoff, J., Zhou, S., Garg, P., Dwivedi, K., Alandejani, F.,
  Gosling, R., Schobs, L., Brook, M., Shahin, Y., et~al.: Machine learning
  cardiac-{MRI} features predict mortality in newly diagnosed pulmonary
  arterial hypertension. European Heart Journal-Digital Health  \textbf{3}(2),
  265--275 (2022)

\bibitem{assadi2022role}
Assadi, H., Alabed, S., Maiter, A., Salehi, M., Li, R., Ripley, D.P., Van~der
  Geest, R.J., Zhong, Y., Zhong, L., Swift, A.J., et~al.: The role of
  artificial intelligence in predicting outcomes by cardiovascular magnetic
  resonance: a comprehensive systematic review. Medicina  \textbf{58}(8), ~1087
  (2022)

\bibitem{emdin2009old}
Emdin, M., Vittorini, S., Passino, C., Clerico, A.: Old and new biomarkers of
  heart failure. European Journal of Heart Failure  \textbf{11}(4),  331--335
  (2009)

\bibitem{garg2022cardiac}
Garg, P., Gosling, R., Swoboda, P., Jones, R., Rothman, A., Wild, J.M., Kiely,
  D.G., Condliffe, R., Alabed, S., Swift, A.J.: Cardiac magnetic resonance
  identifies raised left ventricular filling pressure: prognostic implications.
  European Heart Journal  \textbf{43}(26),  2511--2522 (2022)

\bibitem{huang2020multimodal}
Huang, S.C., Pareek, A., Zamanian, R., Banerjee, I., Lungren, M.P.: Multimodal
  fusion with deep neural networks for leveraging ct imaging and electronic
  health record: a case-study in pulmonary embolism detection. Scientific
  Reports  \textbf{10}(1), ~1--9 (2020)

\bibitem{jain2005score}
Jain, A., Nandakumar, K., Ross, A.: Score normalization in multimodal biometric
  systems. Pattern Recognition  \textbf{38}(12),  2270--2285 (2005)

\bibitem{li2018feature}
Li, J., Cheng, K., Wang, S., Morstatter, F., Trevino, R.P., Tang, J., Liu, H.:
  Feature selection: A data perspective. ACM Computing Surveys (CSUR)
  \textbf{50}(6), ~94 (2018)

\bibitem{lu2022pykale}
Lu, H., Liu, X., Zhou, S., Turner, R., Bai, P., Koot, R.E., Chasmai, M.,
  Schobs, L., Xu, H.: Pykale: Knowledge-aware machine learning from multiple
  sources in python. In: Proceedings of the 31st ACM International Conference
  on Information \& Knowledge Management. pp. 4274--4278 (2022)

\bibitem{lu2013multilinear}
Lu, H., Plataniotis, K.N., Venetsanopoulos, A.: Multilinear subspace learning:
  dimensionality reduction of multidimensional data. CRC press (2013)

\bibitem{lu2008mpca}
Lu, H., Plataniotis, K.N., Venetsanopoulos, A.N.: {MPCA}: Multilinear principal
  component analysis of tensor objects. IEEE Transactions on Neural Networks
  \textbf{19}(1),  18--39 (2008)

\bibitem{scikit-learn}
Pedregosa, F., Varoquaux, G., Gramfort, A., Michel, V., Thirion, B., Grisel,
  O., Blondel, M., Prettenhofer, P., Weiss, R., Dubourg, V., Vanderplas, J.,
  Passos, A., Cournapeau, D., Brucher, M., Perrot, M., Duchesnay, E.:
  Scikit-learn: Machine learning in {P}ython. Journal of Machine Learning
  Research  \textbf{12},  2825--2830 (2011)

\bibitem{sadatsafavi2021moving}
Sadatsafavi, M., Adibi, A., Puhan, M., Gershon, A., Aaron, S.D., Sin, D.D.:
  Moving beyond {AUC}: decision curve analysis for quantifying net benefit of
  risk prediction models. European Respiratory Journal  \textbf{58}(5) (2021)

\bibitem{savarese2022global}
Savarese, G., Becher, P.M., Lund, L.H., Seferovic, P., Rosano, G.M., Coats,
  A.J.: Global burden of heart failure: a comprehensive and updated review of
  epidemiology. Cardiovascular Research  \textbf{118}(17),  3272--3287 (2022)

\bibitem{schobs2022uncertainty}
Sch{\"o}bs, L., Swift, A.J., Lu, H.: Uncertainty estimation for heatmap-based
  landmark localization. IEEE Transactions on Medical Imaging  (2022)

\bibitem{schobs2021confidence}
Schobs, L., Zhou, S., Cogliano, M., Swift, A.J., Lu, H.: Confidence-quantifying
  landmark localisation for cardiac {MRI}. In: 2021 IEEE 18th International
  Symposium on Biomedical Imaging (ISBI). pp. 985--988. IEEE (2021)

\bibitem{swift2021machine}
Swift, A.J., Lu, H., Uthoff, J., Garg, P., Cogliano, M., Taylor, J., Metherall,
  P., Zhou, S., Johns, C.S., Alabed, S., et~al.: A machine learning cardiac
  magnetic resonance approach to extract disease features and automate
  pulmonary arterial hypertension diagnosis. European Heart
  Journal-Cardiovascular Imaging  \textbf{22}(2),  236--245 (2021)

\bibitem{uthoff2020geodesically}
Uthoff, J., Alabed, S., Swift, A.J., Lu, H.: Geodesically smoothed tensor
  features for pulmonary hypertension prognosis using the heart and surrounding
  tissues. In: 23rd International Conference Medical Image Computing and
  Computer Assisted Intervention--MICCAI 2020. pp. 253--262 (2020)

\bibitem{vickers2006decision}
Vickers, A.J., Elkin, E.B.: Decision curve analysis: a novel method for
  evaluating prediction models. Medical Decision Making  \textbf{26}(6),
  565--574 (2006)

\bibitem{welch1947generalization}
Welch, B.L.: The generalization of ‘student's’problem when several
  different population varlances are involved. Biometrika  \textbf{34}(1-2),
  28--35 (1947)

\end{thebibliography}

\end{document}